\documentclass[sigconf]{acmart}

\copyrightyear{2026}
\acmYear{2026}
\setcopyright{cc}
\setcctype{by}
\acmConference[KDD '26]{Proceedings of the 32nd ACM SIGKDD Conference on Knowledge Discovery and Data Mining V.2}{August 09--13, 2026}{Jeju Island, Republic of Korea}
\acmBooktitle{Proceedings of the 32nd ACM SIGKDD Conference on Knowledge Discovery and Data Mining V.2 (KDD '26), August 09--13, 2026, Jeju Island, Republic of Korea}
\acmDOI{10.1145/3770855.3818916}
\acmISBN{979-8-4007-2259-2/2026/08}

\usepackage{graphicx}
\usepackage{booktabs}
\usepackage{array}
\usepackage{multirow}
\usepackage{amsmath}
\usepackage[capitalise,noabbrev]{cleveref}
\usepackage{tikz}
\usepackage{pgfplots}
\usepackage{xspace}
\usetikzlibrary{positioning,arrows.meta,fit,backgrounds,matrix,pgfplots.groupplots}
\pgfplotsset{compat=1.18}

\makeatletter
\let\laplace\@undefined
\makeatother

\usepackage{amsmath,amsfonts,bm}

\def\eqref#1{equation~\ref{#1}}

\def\1{\bm{1}}

\DeclareMathAlphabet{\mathsfit}{\encodingdefault}{\sfdefault}{m}{sl}
\SetMathAlphabet{\mathsfit}{bold}{\encodingdefault}{\sfdefault}{bx}{n}

\newcommand{\laplace}{\mathrm{Laplace}} 

\newcommand{\s}[1]{\textcolor{gray}{\scriptsize{$\pm #1$}}}
\newcommand{\sv}[2]{\s{#1}$^{#2}$}

\newcommand{\ours}{StageGuard\xspace}

\title{StageGuard: Physiologically Constrained Sleep Staging}

\author{Juntang Wang}
\authornote{These authors contributed equally to this work.}
\affiliation{%
  \institution{Zu Chongzhi Center,\\ Duke Kunshan University}%
  \city{Kunshan}\country{China}}
\email{jw853@duke.edu}

\author{Yihan Wang}
\authornotemark[1]
\affiliation{%
  \institution{Zu Chongzhi Center,\\ Duke Kunshan University}%
  \city{Kunshan}\country{China}}
\email{yw572@duke.edu}

\author{Hao Wu}
\authornotemark[1]
\affiliation{%
  \institution{Zu Chongzhi Center,\\ Duke Kunshan University}%
  \city{Kunshan}\country{China}}
\email{hw412@duke.edu}

\author{Jiayu Gao}
\authornotemark[1]
\affiliation{%
  \institution{Zu Chongzhi Center,\\ Duke Kunshan University}%
  \city{Kunshan}\country{China}}
\email{jg563@duke.edu}

\author{Shixin Xu}
\affiliation{%
  \institution{Zu Chongzhi Center,\\ Duke Kunshan University}%
  \city{Kunshan}\country{China}}
\email{shixin.xu@duke.edu}

\author{Dongmian Zou}
\authornote{Corresponding author.}
\affiliation{%
  \institution{Zu Chongzhi Center,\\ Duke Kunshan University}%
  \city{Kunshan}\country{China}}
\email{dongmian.zou@duke.edu}

\renewcommand{\shortauthors}{Juntang Wang et al.}

\begin{abstract}
Automated sleep staging is increasingly used in large-scale studies and downstream scientific analyses to derive sleep-architecture endpoints, including total sleep time, REM latency, sleep efficiency, and bout-duration statistics.
Deep learning models achieve epoch-level accuracy approaching inter-rater agreement, yet often produce hypnograms that violate physiological invariants, such as rare state transitions (e.g., direct Wake$\to$REM) or excessively fragmented sequences.
Such violations can bias downstream sleep metrics, regardless of overall accuracy.
We propose \emph{\ours}, a plug-and-play, backbone-agnostic structured-inference framework that wraps \emph{any} neural sleep-staging backbone with physiology-informed priors.
\ours combines (1) a differentiable soft transition penalty that discourages physiologically rare transitions during training, and (2) a semi-Markov constrained decoder with a duration-augmented state space that jointly enforces transition penalties and minimum bout durations at inference.
Unlike hard-prohibition methods, it admits rare transitions when emission evidence is overwhelming, leaving informative pathological events recoverable rather than blocked.
\ours is a reliability-enabling layer rather than a generative model of sleep; it constrains AI staging outputs to satisfy known physiological priors, ensuring downstream scientific analyses are more trustworthy.
We quantify the validity gap using transition-violation rate (TVR) and fragmentation index (FI) and demonstrate that, across six backbones and four datasets, \ours reduces TVR to physiologically plausible levels and lowers FI by 56--62\%, while maintaining or slightly improving classification accuracy.
Crucially, improved constraint satisfaction translates into 59--79\% lower error on derived sleep-architecture statistics that are \emph{not} directly optimized by the method, and recovers the direction and effect size of expert-defined subgroup differences (OSA severity, age) more faithfully than the unconstrained baseline.
\end{abstract}



\begin{document}
\maketitle

\section{Introduction}\label{sec:intro}

Automated sleep staging---the classification of physiological signals into  vigilance states---is a cornerstone of modern sleep research (e.g., classic R\&K-style standards \citep{rechtschaffenManualStandardizedTerminology1968,wolpertManualStandardizedTerminology1969}).
Sleep can be assessed from diverse signal modalities: polysomnographic EEG/EMG \citep{bargerRobustAutomatedSleep2019}, wrist-worn actigraphy \citep{sadehActivityBasedSleepWakeIdentification1994,vanheesNovelOpenAccess2015}, cardiorespiratory recordings \citep{penzelSystematicComparisonDifferent2002,fonsecaSleepStageClassification2015}, and contactless radar \citep{tataraidzeSleepStageClassification2015}.
Manual scoring by trained experts remains the gold standard, but it is time-consuming, subjective \citep{danker-hopfeInterraterReliabilitySleep2009}, and poorly suited to large-scale studies \citep{guillotDreemOpenDatasets2020}.
Deep learning methods have achieved expert-level accuracy on standard benchmarks \citep{supratakDeepSleepNetModelAutomatic2017,phanSeqSleepNetEndtoEndHierarchical2019,perslevUSleepResilientHighfrequency2021}, leading many laboratories to adopt them for routine scoring.

However, accuracy alone is an incomplete measure of a staging model's fitness for scientific use.
A classifier that achieves high epoch-level accuracy---approaching the inter-rater agreement ceiling of $\sim$80--85\% \citep{danker-hopfeInterraterReliabilitySleep2009,guillotDreemOpenDatasets2020}---may nonetheless produce outputs that violate fundamental physiological invariants: atypical state transitions that occur almost exclusively in pathological conditions (e.g., direct Wake~$\to$~REM in narcolepsy), single-epoch bouts that no sleep researcher would accept, and fragmentation patterns inconsistent with the known temporal organization of sleep \citep{frankenSleepDeprivationRats1991,limSleepFragmentationRisk2013}.
Such outputs are not merely inaccurate---they are \emph{invalid} for downstream analyses including REM latency estimation, sleep efficiency calculation, and bout-duration statistics.
We formalize this validity gap through two constraint-satisfaction indicators: the \emph{transition-violation rate} (TVR), measuring the proportion of physiologically rare transitions, and the \emph{fragmentation index} (FI), measuring excessive state switching.
Our goal is not to optimize these metrics competitively but to ensure they are satisfied at physiologically plausible levels, enabling trustworthy downstream analysis.

The core problem is that standard neural network classifiers treat each epoch independently or with only weak sequential coupling.
Even recurrent and attention-based architectures \citep{phanSeqSleepNetEndtoEndHierarchical2019,eldeleAttentionBasedDeepLearning2021} learn temporal dependencies from data rather than encoding physiological invariants explicitly, leaving them free to produce rare transitions when the learned statistics are ambiguous.
Post-hoc smoothing heuristics (e.g., majority filtering \citep{stephansenNeuralNetworkAnalysis2018}) reduce noise but cannot guarantee that outputs respect known physiological constraints.

As sleep-staging models enter routine use in large-scale studies, outputs that achieve high accuracy but violate physiological invariants can silently corrupt downstream analyses.
Trustworthy deployment requires that outputs satisfy domain-specific constraints, not merely optimize classification metrics \citep{rudinStopExplainingBlack2019}.
To address this gap, we propose \emph{\ours}, a backbone-agnostic structured-inference layer that wraps any neural sleep-staging backbone with physiology-informed priors.
We position \ours as a \emph{reliability-enabling framework} for scientific use of automated staging: it does not propose a new model of sleep biology, but instead operationalizes already-established physiological knowledge as priors over AI outputs so that downstream sleep-architecture analyses derived from automated hypnograms become more trustworthy.

Our contributions are:
\begin{enumerate}
    \item We formalize \emph{transition-violation rate} (TVR) and \emph{fragmentation index} (FI) as validity indicators that complement epoch-level accuracy, and document how unconstrained AI staging biases downstream sleep-architecture endpoints.
    \item We propose a unified, backbone-agnostic framework that combines a soft transition-penalty regularizer during training with a duration-augmented semi-Markov constrained decoder at inference. We explicitly delineate which components admit \emph{exact} structured inference and which are \emph{approximate} or \emph{heuristic} (\Cref{sec:method:scope}).
    \item We demonstrate across six backbones and four datasets spanning distinct modalities that \ours reduces TVR to physiologically plausible levels while maintaining or improving accuracy, reduces downstream sleep-architecture errors by 59--79\%, and better recovers expert-defined subgroup differences in OSA severity and age.
    
\end{enumerate}

\section{Background and Related Work}\label{sec:related}

\subsection{Sleep Physiology and Staging}\label{sec:related:physiology}

Mammalian sleep is organized into cyclically alternating vigilance states: wakefulness (Wake), non-rapid-eye-movement sleep (NREM), and rapid-eye-movement sleep (REM) \citep{rechtschaffenManualStandardizedTerminology1968,wolpertManualStandardizedTerminology1969}.
Transitions between these states follow well-characterized physiological patterns.
The dominant transition pathway is Wake $\to$ NREM $\to$ REM $\to$ Wake; direct REM $\to$ NREM transitions are \emph{rare} under normal conditions. Direct Wake $\to$ REM transitions---termed sleep-onset REM periods (SOREMPs)---rarely occur outside pathological states such as narcolepsy or conditions of extreme sleep deprivation \citep{frankenSleepDeprivationRats1991}.
We emphasize that these transitions are not strictly \emph{forbidden}---they are physiologically \emph{rare} in healthy subjects.
This distinction is important: a staging model should strongly discourage rare transitions but allow them when emission evidence is overwhelming, rather than prohibiting them entirely.

Beyond transition structure, sleep states exhibit characteristic \emph{temporal organization}: bouts of each state have minimum expected durations that reflect the underlying neurobiology of state-switching circuits \citep{saperHypothalamicRegulationSleep2005,luppiNeuroanatomicalNeurochemicalBases2018}.
Sleep fragmentation---excessively short or frequently interrupted bouts---is itself a clinically meaningful outcome associated with cognitive impairment and daytime dysfunction \citep{limSleepFragmentationRisk2013,wolkSleepCardiovascularDisease2005}.
A sleep-staging algorithm that produces unrealistically fragmented output is therefore not only inaccurate but potentially misleading for downstream analyses.

\subsection{Deep Learning for Sleep Staging}\label{sec:related:dl}

Deep learning has achieved expert-level accuracy on sleep staging across modalities.
For polysomnographic EEG, architectures range from CNNs \citep{tsinalisAutomaticSleepStage2016,sorsConvolutionalNeuralNetwork2018,bargerRobustAutomatedSleep2019} to CNN--RNN hybrids \citep{supratakDeepSleepNetModelAutomatic2017,phanSeqSleepNetEndtoEndHierarchical2019}, attention-based models \citep{eldeleAttentionBasedDeepLearning2021,phanSleepTransformerAutomaticSleep2022}, and U-Net architectures \citep{perslevUSleepResilientHighfrequency2021}; see \citet{fiorilloAutomatedSleepScoring2019} for a comprehensive review.
Actigraphy-based sleep--wake classification \citep{sadehActivityBasedSleepWakeIdentification1994,walchSleepStagePrediction2019} and cardiorespiratory staging \citep{fonsecaSleepStageClassification2015} extend these methods to wearable and home-monitoring contexts.
Contactless radar-based methods \citep{tataraidzeSleepStageClassification2015} extend sleep staging to fully touch-free settings by analyzing respiratory and movement patterns from reflected radio waves.
However, despite this progress, these deep learning methods do not explicitly encode physiological transition constraints.
Sequential models learn soft temporal dependencies but provide no mechanism to ensure outputs respect known physiological invariants; rare transitions and excessive fragmentation can bias downstream sleep architecture statistics.

Recent work on foundation models for EEG \citep{jiangLargeBrainModel2024} demonstrates the potential of large-scale pre-training across diverse EEG tasks.
Such models learn rich representations but do not explicitly encode physiological constraints.
Our framework is orthogonal and complementary: \ours can wrap any backbone, including fine-tuned foundation models, to ensure outputs respect domain-specific invariants regardless of representation quality.
To incorporate such domain knowledge, we review relevant work on constraints in machine learning.

\subsection{Constraints in Machine Learning}\label{sec:related:constraints}

Incorporating domain knowledge as constraints in machine learning dates to early work on graphical models and structured prediction.
In structured prediction, conditional random fields \citep{laffertyConditionalRandomFields2001} and hidden Markov models (HMMs) encode transition structure via learned or fixed transition matrices \citep{koleyEnsembleSystemAutomatic2012}.
The Viterbi algorithm \citep{forneyViterbiAlgorithm1973} finds the most likely state sequence under such models and can incorporate transition constraints via the log-transition matrix.

A key limitation of HMMs is their assumption of geometric (memoryless) state durations.
Hidden semi-Markov models (HSMMs) \citep{yuHiddenSemiMarkovModels2010} generalize HMMs by explicitly modeling state duration distributions, enabling minimum dwell-time constraints that better capture the temporal structure of physiological processes.
Our minimum-duration constraint mechanism approximates semi-Markov behavior within a computationally efficient Viterbi framework.

In natural language processing, constrained decoding has been a powerful technique for incorporating lexical and structural constraints into neural sequence generation.
Grid beam search \citep{hokampLexicallyConstrainedDecoding2017} and dynamic beam allocation \citep{postFastLexicallyConstrained2018} enable hard lexical constraints during decoding while preserving model fluency.
Our constrained Viterbi decoder applies an analogous principle---enforcing domain-specific structural constraints on neural network outputs---to the sequential classification setting.
In sleep staging,
\citet{dongMixedNeuralNetwork2018} used a mixed neural network with CRF-like post-processing, and hidden Markov models have been applied to sleep state sequences \citep{panTransitionconstrainedDiscreteHidden2012}. However, neither approach provides a backbone-agnostic wrapper with semi-Markov inference.
Our work is conceptually related to physics-informed machine learning \citep{karpatneTheoryGuidedDataScience2017,willardIntegratingScientificKnowledge2023}, though we encode discrete physiological rules rather than governing equations.

\paragraph{Summary and gap.}
No existing framework combines soft training penalties with principled semi-Markov inference into a unified, backbone-agnostic wrapper for physiological time-series classification.
Our work addresses this gap by integrating constraints---soft transition penalties during training, duration-augmented semi-Markov decoding at inference---that can augment any neural sleep-staging backbone without architectural modification.

\section{Method}\label{sec:method}

We consider sleep staging as a structured sequence prediction problem.
Given an epoch sequence of inputs $\mathbf{x} = (x_1, \ldots, x_T)$ (with modality-specific representations), a backbone model produces per-epoch posterior distributions over sleep states:
$p_{\theta}(y_t \mid x_t)$, where $y_t \in \mathcal{S}$.
Our goal is to output a hypnogram $\mathbf{y} = (y_1, \ldots, y_T)$ that is not only accurate but also physiologically plausible.
In particular, we target two common failure modes of modern deep sleep-staging systems: (i)~\emph{transition violations} and (ii)~\emph{temporal fragmentation}.
\ours is a plug-and-play, backbone-agnostic inference layer that turns any per-epoch classifier into a physiologically consistent scientific readout.

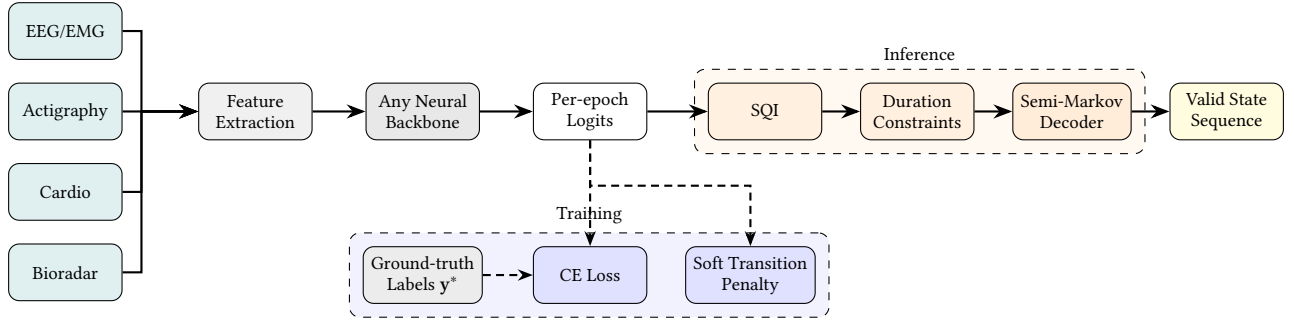
\begin{figure*}[t]
    \centering
    \begin{tikzpicture}[
    node distance=0.7cm and 1.0cm,
    box/.style={rectangle, draw, rounded corners, minimum height=0.75cm, minimum width=1.5cm, align=center, fill=white, font=\footnotesize},
    dashed box/.style={rectangle, draw, dashed, rounded corners, inner sep=5pt},
    arr/.style={-{Stealth[length=2.5mm]}, thick},
    darr/.style={-{Stealth[length=2mm]}, thick, densely dashed},
    every node/.style={font=\small}
]
    \node[box, fill=teal!12] (eeg) {EEG/EMG};
    \node[box, fill=teal!12, below=0.3cm of eeg] (acti) {Actigraphy};
    \node[box, fill=teal!12, below=0.3cm of acti] (cardio) {Cardio};
    \node[box, fill=teal!12, below=0.3cm of cardio] (radar) {Bioradar};

    \node[box, fill=gray!12, right=1.0cm of acti] (feat) {Feature\\Extraction};

    \node[box, fill=gray!18, right=0.7cm of feat] (cnn) {Any Neural\\Backbone};

    \node[box, right=0.7cm of cnn] (logits) {Per-epoch\\Logits};

    \node[box, fill=gray!15, below=1.4cm of cnn] (gtlabels) {Ground-truth\\Labels $\mathbf{y}^*$};

    \node[box, fill=blue!12, below=1.4cm of logits] (celoss) {CE Loss};
    \node[box, fill=blue!12, right=0.5cm of celoss] (softpen) {Soft Transition\\Penalty};

    \node[box, fill=orange!12, right=0.8cm of logits] (sqi) {SQI};
    \node[box, fill=orange!12, right=0.5cm of sqi] (temporal) {Duration\\Constraints};
    \node[box, fill=orange!15, right=0.5cm of temporal] (viterbi) {Semi-Markov\\Decoder};
    \node[box, fill=yellow!15, right=0.5cm of viterbi] (output) {Valid State\\Sequence};

    \draw[arr] (eeg.east) -- ++(0.25,0) |- (feat.west);
    \draw[arr] (acti.east) -- (feat.west);
    \draw[arr] (cardio.east) -- ++(0.25,0) |- (feat.west);
    \draw[arr] (radar.east) -- ++(0.25,0) |- (feat.west);

    \draw[arr] (feat) -- (cnn);
    \draw[arr] (cnn) -- (logits);

    \draw[arr] (logits) -- (sqi);
    \draw[arr] (sqi) -- (temporal);
    \draw[arr] (temporal) -- (viterbi);
    \draw[arr] (viterbi) -- (output);

    \draw[darr] (logits) -- (celoss);
    \draw[darr] (logits.south) -- ++(0,-0.6) -| (softpen.north);
    \draw[darr] (gtlabels) -- (celoss);

    \begin{scope}[on background layer]
        \node[dashed box, fill=blue!5, fit=(celoss)(softpen)(gtlabels), label={[font=\footnotesize]above:Training}] {};
        \node[dashed box, fill=orange!5, fit=(sqi)(temporal)(viterbi), label={[font=\footnotesize]above:Inference}] {};
    \end{scope}
\end{tikzpicture}
    \caption{Overview of the \ours pipeline. Input features from one modality (EEG/EMG, actigraphy, cardiorespiratory, or bioradar signals) are processed by a modality-specific feature extraction stage and \emph{any} neural backbone to produce per-epoch logits. During training (blue box, dashed arrows), cross-entropy loss computed against ground-truth labels $\mathbf{y}^*$ is augmented with a soft transition penalty that discourages rare transitions. At inference (orange box), a signal-quality index (SQI) module attenuates unreliable emissions (Eq. \ref{eq:sqi}), followed by duration constraints and a semi-Markov decoder that jointly enforce transition constraints and minimum bout durations, producing a physiologically valid state sequence. All constraint modules are backbone-agnostic and operate as a plug-and-play wrapper.}
    \Description{Schematic pipeline diagram read left to right. Input signals from a single modality (EEG/EMG, actigraphy, cardiorespiratory, or bioradar) pass through a modality-specific feature-extraction stage and then any neural backbone, which outputs per-epoch class logits. A dashed blue training path adds a soft transition penalty to the cross-entropy loss to discourage physiologically rare transitions. A solid orange inference path sends the logits through a signal-quality module that down-weights unreliable epochs, then duration constraints and a semi-Markov decoder that jointly enforce transition penalties and minimum bout durations, yielding a physiologically valid sleep-stage sequence. All constraint modules sit outside the backbone as a plug-and-play wrapper.}\label{fig:pipeline}
\end{figure*}

\subsection{Problem Formulation}\label{sec:method:formulation}

Let $\mathbf{x} = (x_1, x_2, \ldots, x_T)$ denote a sequence of $T$ feature vectors extracted from consecutive epochs of any physiological signal, where $T$ is the recording length in epochs, each $x_t \in \mathbb{R}^{D_f}$, and $D_f$ depends on the modality.
The goal is to predict a label sequence $\mathbf{y} = (y_1, y_2, \ldots, y_T)$ with $y_t \in \mathcal{S}$, where $\mathcal{S}$ is a modality-appropriate state set (e.g., $\{\text{Wake}, \text{NREM}, \text{REM}\}$ for EEG/EMG, $\{\text{Wake}, \text{Sleep}\}$ for actigraphy) with $K = |\mathcal{S}|$ states.
A neural backbone with parameters $\theta$ maps each epoch to a posterior distribution $p_\theta(y_t \mid x_t)$ over $\mathcal{S}$.
This is a \emph{structured output prediction} problem: rather than classifying each epoch independently, the output sequence must satisfy global structural constraints imposed by sleep physiology.

We define a \emph{rare-transition} indicator for adjacent epochs:
\begin{equation}\label{eq:violation}
    v_t = \mathbf{1}\bigl[(y_{t-1}, y_t) \in \mathcal{R}\bigr], \quad t = 2, \ldots, T,
\end{equation}
where $\mathcal{R} \subset \mathcal{S} \times \mathcal{S}$ is the set of physiologically \emph{rare} transitions (e.g., Wake$\to$REM, REM$\to$NREM in healthy subjects); see \Cref{tab:rare} for modality-specific definitions.
The \emph{transition-violation rate} (TVR) measures the proportion of rare transitions: $\text{TVR}(\mathbf{y}) = \frac{1}{T-1}\sum_{t=2}^{T} v_t$.

\begin{table}[t]
\caption{Physiologically rare transitions $\mathcal{R}$ by modality. W=Wake, R=REM, N=NREM.}\label{tab:rare}
\centering
\small
\begin{tabular}{lll}
\toprule
Modality & Rare Transitions & Rationale \\
\midrule
EEG/EMG (3-state) & W$\to$R, R$\to$N & SOREMPs pathological \\
Cardiorespiratory & W$\to$R, R$\to$N & Same \\
Bioradar (3-state) & W$\to$R, R$\to$N & Same \\
Actigraphy (2-state) & --- & None \\
\bottomrule
\end{tabular}
\end{table}
We emphasize that TVR is a \emph{validity indicator}, not a performance metric: a model with high TVR produces outputs inconsistent with normal sleep physiology, regardless of accuracy.

We additionally define a \emph{fragmentation index} to capture temporal consistency:
\begin{equation}\label{eq:frag}
    \text{FI}(\mathbf{y}) = \frac{1}{T - 1}\sum_{t=2}^{T} \mathbf{1}[y_{t-1} \neq y_t].
\end{equation}
A high FI indicates excessive state switching relative to the recording length.
We also track mean bout duration (the average length of contiguous runs of the same state) as a complementary temporal consistency metric.

\subsection{Backbone-Agnostic Design}\label{sec:method:cnn}

\ours is designed as a plug-and-play constraint module that wraps any neural backbone producing per-epoch class probabilities $p_\theta(y_t \mid x_t) \in \Delta^{|\mathcal{S}|-1}$ (the probability simplex).
The only requirement is that the backbone outputs a softmax distribution over states for each epoch; no architectural modifications are needed.

We evaluate six established backbones: AccuSleep \citep{bargerRobustAutomatedSleep2019}, DeepSleepNet \citep{supratakDeepSleepNetModelAutomatic2017}, SeqSleepNet \citep{phanSeqSleepNetEndtoEndHierarchical2019}, AttnSleep \citep{eldeleAttentionBasedDeepLearning2021}, SleepTransformer \citep{phanSleepTransformerAutomaticSleep2022}, and U-Sleep \citep{perslevUSleepResilientHighfrequency2021}.
These span CNN, RNN, attention, and transformer architectures.

The input representation $x_t$ is modality-specific: for EEG/EMG, $x_t$ is a concatenated spectrogram \citep{bargerRobustAutomatedSleep2019}; for actigraphy, $x_t$ comprises activity count features and circadian covariates \citep{sadehActivityBasedSleepWakeIdentification1994}; for cardiorespiratory data, $x_t$ consists of heart rate variability and respiratory features \citep{fonsecaSleepStageClassification2015}; for bioradar, $x_t$ consists of respiratory amplitude, rate, and body movement features extracted from radar phase signals \citep{tataraidzeSleepStageClassification2015}.
Each backbone is trained with cross-entropy loss:
\begin{equation}\label{eq:ce}
    \mathcal{L}_{\text{CE}} = -\frac{1}{T}\sum_{t=1}^{T} \log p_\theta(y_t^* \mid x_t),
\end{equation}
where $y_t^*$ is the expert label. Throughout, $\log$ denotes the natural logarithm.

\subsection{Soft Transition-Penalty Regularization}\label{sec:method:soft}

Constrained decoding alone cannot improve the backbone's learned representations.
We therefore augment the training loss with a differentiable penalty that steers the backbone toward producing fewer rare transitions in the first place.
For each pair of consecutive epochs $(t-1, t)$, we compute the probability that the model assigns to a physiologically rare transition:
\begin{equation}\label{eq:soft}
    \mathcal{L}_{\text{trans}} = \frac{1}{T-1}\sum_{t=2}^{T} \sum_{(s,s') \in \mathcal{R}} p_\theta(y_{t-1}=s \mid x_{t-1})\, p_\theta(y_t=s' \mid x_t).
\end{equation}
Note that the penalty is not normalized by $|\mathcal{R}|$; the weight $\lambda$ is calibrated accordingly.
Moreover, \Cref{eq:soft} approximates the joint rare-transition probability as a product of per-epoch marginals; for backbones with sequential context (e.g., recurrent or attention-based models), these marginals are not strictly independent. In practice, the penalty serves as a regularizer rather than an exact probability estimate, and empirically yields consistent improvements across all tested architectures regardless of their temporal modeling capacity.
The total training loss becomes:
\begin{equation}\label{eq:total_loss}
    \mathcal{L} = \mathcal{L}_{\text{CE}} + \lambda\, \mathcal{L}_{\text{trans}},
\end{equation}
where $\lambda \geq 0$ controls the penalty strength.
This formulation is related to posterior regularization \citep{ganchevPosteriorRegularizationStructured2010} and encourages the backbone to learn representations that naturally respect physiological constraints.

\begin{table*}[t]
\caption{Exact, approximate, and heuristic components of \ours. The augmented-state decoder is exact under its stated static objective, while the soft penalty and flip-flop term are approximate and heuristic refinements.}\label{tab:scope}
\centering
\begin{tabular}{@{}ll>{\raggedright\arraybackslash}p{10.5cm}@{}}
\toprule
Component & Type & Justification \\
\midrule
Semi-Markov Viterbi decoder (Eq.~\ref{eq:viterbi}) & Exact & MAP inference on augmented state space; recovers globally optimal path under the static transition/duration scores. \\
Static transition constraints (Eq.~\ref{eq:aug_trans}) & Exact & Log-probability penalties from empirical transition matrices (\Cref{tab:empirical_trans}); enforced exactly inside Viterbi. \\
Duration constraints ($d_{\min}$, \S\ref{sec:method:temporal}) & Exact & Hard minimum dwell time enforced exactly via the augmented state space. \\
Soft transition penalty (Eq.~\ref{eq:soft}) & Approximate & Per-epoch marginal product used as a tractable proxy for joint rare-transition probability; acts as a training-time regularizer rather than an exact probability estimate. \\
Flip-flop penalty (Eq.~\ref{eq:flipflop}) & Heuristic & Path-dependent term added greedily during decoding; empirically alters at least one decoded transition in only ${\sim}3.8\%$ of recordings, mostly affecting FI (\Cref{tab:flipflop_ablation}). \\
\bottomrule
\end{tabular}
\end{table*}

\subsection{Semi-Markov Constrained Decoding}\label{sec:method:viterbi}

While the soft penalty steers the model toward plausible predictions, it cannot guarantee that outputs respect physiological constraints.
We apply a semi-Markov constrained decoder at inference time that jointly enforces transition constraints and minimum bout durations within a unified framework.

\paragraph{State augmentation.}
Standard HMMs assume geometric (memoryless) state durations, which is violated in sleep physiology \citep{saperHypothalamicRegulationSleep2005}.
We augment the state space to $\tilde{\mathcal{S}} = \{(s, d) : s \in \mathcal{S}, d \in \{1, \ldots, D_{\max}\}\}$ (with $D_{\max} = 10$ in all experiments), where $d$ counts consecutive epochs in state $s$.

\paragraph{Transition constraints in augmented space.}
Transitions between augmented states follow:
\begin{itemize}
    \item $(s, d) \to (s, d+1)$: continue in state $s$, incrementing duration (if $d < D_{\max}$)
    \item $(s, d) \to (s', 1)$: transition to new state $s'$ with $d=1$, allowed only if $d \geq d_{\min}(s)$
\end{itemize}
The log-transition matrix for the augmented space encodes both duration and transition constraints:
\begin{equation}\label{eq:aug_trans}
\tilde{A}_{(s,d),(s',d')} = \begin{cases}
0 & \text{if } s' = s, d' = \min(d+1, D_{\max}) \\
\log \epsilon_{ss'} & \text{if } s' \neq s, d' = 1, d \geq d_{\min}(s) \\
-\infty & \text{otherwise}
\end{cases}
\end{equation}
where $\epsilon_{ss'}$ encodes the physiological rarity of transition $s \to s'$.
For typical transitions (e.g., NREM$\to$REM), $\epsilon_{ss'} = \hat{\pi}_{ss'}$, where $\hat{\pi}_{ss'} = N_{ss'}/\sum_{j} N_{sj}$ is the empirical transition probability estimated from training-set labels (with $N_{ss'}$ denoting the count of observed $s \to s'$ transitions).
For rare transitions (e.g., Wake$\to$REM), we set $\epsilon_{ss'} = 0.001$---chosen to be small relative to typical transition probabilities ($\hat{\pi}_{ss'} \approx 0.1$--$0.3$) while remaining numerically stable---strongly discouraging but not prohibiting them.

\paragraph{Semi-Markov Viterbi inference.}
The Viterbi algorithm on the augmented state space finds:
\begin{equation}\label{eq:viterbi}
    \hat{\tilde{\mathbf{y}}} = \arg\max_{\tilde{\mathbf{y}}} \left[\sum_{t=1}^{T} E_{t,s_t} + \sum_{t=2}^{T} \tilde{A}_{\tilde{y}_{t-1}, \tilde{y}_t}\right],
\end{equation}
where $E_{t,s} = \log p_\theta(y_t = s \mid x_t)$ is the emission log-probability and $\tilde{y}_t = (s_t, d_t)$ is the augmented state at time $t$.
The final decoded sequence extracts the state component: $\hat{y}_t = \hat{s}_t$.
The initial state distribution is uniform over $\{(s, 1) : s \in \mathcal{S}\}$, assigning equal prior probability to each state at duration count $d{=}1$.

This ensures minimum bout durations are enforced exactly within decoding, and rare transitions are strongly penalized but allowed when emission evidence is overwhelming (i.e., the emission log-probability outweighs the transition penalty).
Inference complexity is $O(T \times (K \cdot D_{\max})^2)$; for $K=3$ and $D_{\max}=10$ (sufficient to capture durations up to 40 seconds for EEG/EMG or 5 minutes for PSG, encompassing typical bout lengths), this yields 30 augmented states but remains acceptable for batch processing.
For context, the backbone forward pass is $O(T \times C)$ where $C$ denotes the backbone's per-epoch forward-pass cost (ranging from ${\sim}10^5$ to ${\sim}10^7$ multiply-accumulate operations across our backbones); the decoder overhead of $O(T \times 900)$ for $K = 3$, $D_{\max} = 10$ is thus negligible for larger backbones.

\subsection{Duration Constraints}\label{sec:method:temporal}

Sleep bouts have characteristic minimum durations reflecting neural circuit dynamics \citep{saperHypothalamicRegulationSleep2005,luppiNeuroanatomicalNeurochemicalBases2018}.
Each state $s$ has a minimum bout duration $d_{\min}(s)$ enforced via \Cref{eq:aug_trans}:
\begin{itemize}
    \item \textbf{EEG/EMG (4-sec epochs):} $d_{\min}(\text{NREM}) = 3$, $d_{\min}(\text{REM}) = 2$, $d_{\min}(\text{Wake}) = 2$
    \item \textbf{PSG (30-sec epochs):} $d_{\min}(\text{NREM}) = 2$, $d_{\min}(\text{REM}) = 2$, $d_{\min}(\text{Wake}) = 1$
    \item \textbf{Bioradar (30-sec epochs):} $d_{\min}(\text{NREM}) = 2$, $d_{\min}(\text{REM}) = 2$, $d_{\min}(\text{Wake}) = 1$
    \item \textbf{Actigraphy (30-sec epochs):} $d_{\min}(\text{Sleep}) = 2$, $d_{\min}(\text{Wake}) = 1$
\end{itemize}
These thresholds are informed by established sleep physiology \citep{saperHypothalamicRegulationSleep2005} and empirically validated on held-out folds.

To suppress rapid state alternation (flip-flop), we apply an additional penalty when transitioning to a state visited within the previous $k$ epochs.
Specifically, when proposing a transition from state $s'$ to state $s$ at time $t$:
\begin{equation}\label{eq:flipflop}
    \rho(s', s, t) = -\gamma \cdot \mathbf{1}\bigl[s \neq s' \;\land\; s \in \{y_{t-2}, \ldots, y_{t-k}\}\bigr],
\end{equation}
where $\gamma = 2.0$ and $k = 5$.
This penalty is added to the augmented transition scores in \Cref{eq:aug_trans} at runtime based on the decoded history.
Because the flip-flop penalty depends on the decoded path, it is applied as a greedy augmentation within the Viterbi forward pass; the resulting sequence is therefore optimal with respect to the emission and static transition scores, with the flip-flop penalty providing an additional heuristic refinement that empirically affects fewer than 5\% of decoded transitions.
Unlike post-hoc smoothing, our semi-Markov formulation enforces transition and duration constraints within decoding.

\subsection{Signal-Quality Handling}\label{sec:method:quality}

Real-world recordings contain corrupted epochs due to sensor detachment, motion artifacts, or missing segments.
For each modality, we compute a signal-quality indicator $\beta_t \in [0,1]$ for each epoch, where $\beta_t = 0$ denotes a clean epoch and $\beta_t = 1$ denotes a fully corrupted epoch.
The thresholds below are drawn from established literature- and physiology-based criteria rather than tuned per dataset, and the same thresholds are applied across all four datasets without retuning:
for EEG/EMG, per-epoch amplitude $z$-scores ($|z|>3$) or high-frequency ($>$\,45\,Hz) power fraction exceeding 0.5, following standard polysomnography artifact-rejection criteria \citep{berryAASMManualScoring2017};
for actigraphy, zero activity counts persisting ${\geq}\,3$ consecutive epochs (device-removal detection) following clinical actigraphy guidelines \citep{sadehRoleValidityActigraphy2011};
for cardiorespiratory data, heart rate outside 30--200\,bpm or missing signal segments, consistent with physiological plausibility bounds in HRV analysis \citep{shafferOverviewHeartRate2017};
for bioradar, amplitude $z$-scores ($|z|>5$) or signal dropouts spanning ${\geq}\,2$ consecutive epochs, following sensor-specific noise characteristics of contactless radar.
Detected low-quality epochs have their emission log-probabilities interpolated toward the uniform distribution (reflecting maximum uncertainty when signal quality is compromised):
\begin{equation}\label{eq:sqi}
    \tilde{E}_{t,s} = (1 - \beta_t)\, E_{t,s} + \beta_t\, \log(1/|\mathcal{S}|),
\end{equation}
where $\beta_t$ reflects the degree of signal degradation at epoch $t$.
When signal-quality information is available, the decoder (\Cref{eq:viterbi}) uses $\tilde{E}_{t,s}$ in place of $E_{t,s}$, allowing the transition model to bridge across corrupted segments.

\subsection{Exact, Approximate, and Heuristic Components}\label{sec:method:scope}

To make \ours's guarantees explicit rather than treat it as a black-box wrapper, we classify each component as \emph{exact} (globally optimal under its stated objective), \emph{approximate} (a tractable surrogate for the quantity of interest), or \emph{heuristic} (a hand-designed refinement without an optimality guarantee). \Cref{tab:scope} summarizes this decomposition.  

The validity guarantees of \ours come from the exact augmented-state decoder: under fixed emission scores, transition penalties, and minimum-duration thresholds, the decoded path is globally optimal with respect to the static objective in Eq. \ref{eq:viterbi}.
The soft penalty improves the backbone's logits during training (\Cref{tab:ablation}) but its marginal-product form does not faithfully model the true joint rare-transition probability for sequential backbones, and the flip-flop term changes only ${\sim}3.8\%$ of recordings (\Cref{tab:flipflop_ablation})---both useful refinements, but conceptually separable from the exact core.

\section{Experiments}\label{sec:experiments}

\subsection{Datasets}\label{sec:exp:data}

We evaluate on four publicly available datasets:
\textbf{(1) AccuSleep Mouse EEG/EMG} \citep{bargerRobustAutomatedSleep2019}: 16 mice, 24-hour recordings at 512~Hz, 4-second epochs, three states (Wake, NREM and REM).
\textbf{(2) Sleep-Accel} \citep{walchSleepStagePrediction2019}: 31 adults with wrist actigraphy, 30-second epochs, two states (Wake/Sleep); all transitions allowed, so temporal consistency is the primary constraint mechanism.
\textbf{(3) SHHS} \citep{quanSleepHeartHealth1997}: 25 subjects (stratified random sample by age and OSA severity; selection seed 2024; SHHS-1 visit-1 subject IDs 200001--204000 pool) with cardiorespiratory features (HRV, respiratory rate), 30-second epochs, three states; transition constraints mirror EEG/EMG.
\textbf{(4) SLEEPBRL} \citep{tataraidzeSleepStageClassification2015}: 32 subjects monitored by contactless bioradar (3.6--4.0~GHz continuous-wave) in a sleep laboratory with simultaneous PSG; AASM-scored 30-second epochs collapsed to three states (Wake/NREM/REM). This is the only fully touch-free modality in our evaluation.

\subsection{Baselines}\label{sec:exp:baselines}

We compare six neural backbones, each evaluated both standalone and augmented with the full \ours constraint framework (soft penalty + constrained Viterbi + temporal consistency):
\begin{enumerate}
    \item \textbf{AccuSleep} \citep{bargerRobustAutomatedSleep2019}: compact CNN with two convolutional layers.
    \item \textbf{DeepSleepNet} \citep{supratakDeepSleepNetModelAutomatic2017}: CNN--RNN hybrid with multi-scale filters and bidirectional LSTM.
    \item \textbf{SeqSleepNet} \citep{phanSeqSleepNetEndtoEndHierarchical2019}: hierarchical RNN with attention.
    \item \textbf{AttnSleep} \citep{eldeleAttentionBasedDeepLearning2021}: multi-resolution CNN with multi-head self-attention.
    \item \textbf{SleepTransformer} \citep{phanSleepTransformerAutomaticSleep2022}: transformer-based with epoch- and sequence-level attention.
    \item \textbf{U-Sleep} \citep{perslevUSleepResilientHighfrequency2021}: fully convolutional U-Net trained on large-scale data.
\end{enumerate}

For each dataset, all backbones use the same input features, differing only in architecture.
The ``+Ours'' variant adds the full \ours constraint module with soft penalty weight $\lambda = 0.5$, dataset-specific minimum-duration thresholds $d_{\min}$, flip-flop window $k = 5$ epochs, and flip-flop penalty weight $\gamma = 2.0$.
We evaluate using leave-one-subject-out (LOSO) cross-validation, using one fold per subject/recording: 16 folds for AccuSleep Mouse EEG/EMG, 31 folds for Sleep-Accel, 25 folds for SHHS, and 32 folds for SLEEPBRL.
In each fold, we train on all but one subject and report metrics on the held-out subject; the final ``mean $\pm$ std'' aggregates across folds.

\subsection{Training and Implementation Details}\label{sec:exp:impl}

\paragraph{Training protocol.}
For each backbone, we follow the training recipe recommended in its original work (optimizer, learning-rate schedule, batch size, and early stopping), and we keep the protocol identical between the baseline and ``+Ours'' variants so that differences are attributable to the constraint wrapper rather than additional tuning.
The loss is cross-entropy plus the soft transition penalty (\Cref{eq:total_loss}) for ``+Ours''.

\paragraph{Hyperparameter selection.}
We use a single set of constraint hyperparameters across backbones for fairness.
The soft-penalty weight $\lambda = 0.5$ was selected via grid search over $\{0.1, 0.25, 0.5, 1.0, 2.0\}$ on a held-out validation fold (one subject per dataset), optimizing for TVR reduction while maintaining baseline accuracy; sensitivity analysis is provided in \Cref{tab:lambda}.
The flip-flop window $k{=}5$ epochs corresponds to 20\,s for 4-second epochs and 2.5\,min for 30-second epochs, both within the refractory period following a state transition \citep{saperHypothalamicRegulationSleep2005}; $\gamma{=}2.0$ was selected alongside $\lambda$ on the validation fold. Minimum-duration thresholds $d_{\min}$ are physiology-informed (\Cref{sec:method:temporal}); all values are reported in the released configuration files.

\paragraph{Randomness and variance reporting.}
We compute ``mean $\pm$ std'' across LOSO folds (one fold per held-out subject), using a fixed random seed of 42 for weight initialization, data shuffling, and fold assignment within each fold.
We will release per-fold metrics to enable paired significance testing and confidence-interval estimation.

\paragraph{Compute and runtime.}
All experiments were conducted on a single NVIDIA A100 GPU (40\,GB).
The decision-critical deployment figures are reported directly in the main text (see \emph{Computational overhead} below): the soft penalty adds ${<}1\%$ training overhead, and the constrained decoder adds only a small per-recording inference overhead. The complete per-fold training-time and inference-throughput logs are released with the artifact for full reproducibility.

\paragraph{Computational overhead.}
The semi-Markov decoder adds minimal overhead: for U-Sleep (the largest backbone), mean inference time increases from 12.3\,ms to 13.1\,ms per subject (full recording) (+6.5\%), dominated by the backbone forward pass.
For smaller backbones (AccuSleep), relative overhead is higher (+18\%) but absolute time remains negligible ($<$2\,ms).
Memory overhead is $O(T \cdot K \cdot D_{\max})$ for the augmented state space (Viterbi backpointers), adding a few MB for typical 24-hour recordings.

\subsection{Evaluation Metrics}\label{sec:exp:metrics}

We report the following metrics, averaged across cross-validation folds:
\begin{itemize}
    \item \textbf{TVR} (\%): transition-violation rate---the proportion of rare transitions (\Cref{eq:violation}). We emphasize that TVR is a \emph{validity indicator}: it measures whether outputs respect physiological constraints, not model performance. A non-zero TVR indicates outputs that are inconsistent with normal sleep physiology.
    \item \textbf{FI}: fragmentation index (\Cref{eq:frag}), measuring excessive state switching.
    \item \textbf{Acc} (\%): overall epoch-level accuracy.
    \item \textbf{$\kappa$}: Cohen's kappa agreement coefficient \citep{cohenCoefficientAgreementNominal1960}.
    \item \textbf{Per-class F1}: class-wise F1 scores to characterize which stages benefit most; full results are provided in \Cref{sec:appendix:f1}.
\end{itemize}

\subsection{Main Results}\label{sec:exp:results}

\begin{table*}[t]
\caption{Main results across six backbones and four datasets (mean $\pm$ std across cross-validation folds$^\ddagger$). ``+Ours'' denotes the backbone augmented with the full \ours constraint framework. TVR measures the rate of physiologically rare transitions (validity indicator); a non-zero TVR indicates outputs inconsistent with normal sleep physiology. \textbf{Bold}: best per column within each dataset. Statistical significance of +Ours vs.\ baseline tested via Wilcoxon signed-rank test with Bonferroni correction (correcting within each dataset for 6 backbone comparisons per metric): $^{***}p<0.001$, $^{**}p<0.01$, $^{*}p<0.05$ (all thresholds reflect Bonferroni-corrected $p$-values).}\label{tab:main}
\centering
\footnotesize
\resizebox{\textwidth}{!}{%
\begin{tabular}{l lll lll lll lll}
\toprule
 & \multicolumn{3}{c}{AccuSleep Mouse EEG/EMG} & \multicolumn{3}{c}{Sleep-Accel (Actigraphy)} & \multicolumn{3}{c}{SHHS} & \multicolumn{3}{c}{SLEEPBRL (Bioradar)} \\
\cmidrule(lr){2-4} \cmidrule(lr){5-7} \cmidrule(lr){8-10} \cmidrule(lr){11-13}
Backbone & TVR$\downarrow$ & FI$\downarrow$ & Acc$\uparrow$ & TVR$\downarrow$ & FI$\downarrow$ & Acc$\uparrow$ & TVR$\downarrow$ & FI$\downarrow$ & Acc$\uparrow$ & TVR$\downarrow$ & FI$\downarrow$ & Acc$\uparrow$ \\
\midrule
AccuSleep & $8.7$\s{0.7} & $0.21$\s{0.01} & $89.3$\s{0.7} & --- & $0.25$\s{0.01} & $84.1$\s{0.7} & $8.1$\s{0.6} & $0.22$\s{0.01} & $76.8$\s{0.8} & $12.3$\s{0.9} & $0.28$\s{0.02} & $68.4$\s{1.1} \\
\quad +Ours & $0.6$\sv{0.1}{***} & $0.08$\sv{0.006}{***} & $91.0$\sv{0.6}{**} & --- & $0.10$\sv{0.01}{***} & $87.4$\sv{0.6}{**} & $0.5$\sv{0.1}{***} & $0.09$\sv{0.007}{***} & $80.5$\sv{0.7}{**} & $0.8$\sv{0.2}{***} & $0.12$\sv{0.01}{***} & $73.2$\sv{1.0}{**} \\
\addlinespace
DeepSleepNet & $6.4$\s{0.5} & $0.17$\s{0.01} & $90.4$\s{0.6} & --- & $0.21$\s{0.01} & $85.7$\s{0.6} & $6.0$\s{0.5} & $0.18$\s{0.01} & $78.9$\s{0.7} & $10.1$\s{0.8} & $0.24$\s{0.02} & $70.6$\s{1.0} \\
\quad +Ours & $0.4$\sv{0.1}{***} & $0.07$\sv{0.005}{***} & $91.5$\sv{0.6}{**} & --- & $0.09$\sv{0.006}{***} & $85.6$\sv{0.6}{**} & $0.3$\sv{0.06}{***} & $0.08$\sv{0.005}{***} & $81.3$\sv{0.6}{**} & $0.5$\sv{0.1}{***} & $0.10$\sv{0.01}{***} & $74.5$\sv{0.9}{**} \\
\addlinespace
SeqSleepNet & $5.6$\s{0.5} & $0.16$\s{0.01} & $90.8$\s{0.6} & --- & $0.19$\s{0.01} & $86.2$\s{0.6} & $5.3$\s{0.4} & $0.17$\s{0.01} & $79.4$\s{0.7} & $9.2$\s{0.7} & $0.23$\s{0.01} & $71.4$\s{0.9} \\
\quad +Ours & $0.3$\sv{0.07}{***} & $0.07$\sv{0.004}{***} & $91.7$\sv{0.5}{**} & --- & $0.08$\sv{0.005}{***} & $88.6$\sv{0.5}{**} & $0.3$\sv{0.06}{***} & $0.07$\sv{0.004}{***} & $81.7$\sv{0.6}{**} & $0.5$\sv{0.1}{***} & $0.10$\sv{0.01}{***} & $75.2$\sv{0.8}{**} \\
\addlinespace
AttnSleep & $4.5$\s{0.4} & $0.14$\s{0.01} & $91.1$\s{0.5} & --- & $0.18$\s{0.01} & $86.5$\s{0.5} & $4.2$\s{0.3} & $0.15$\s{0.01} & $79.9$\s{0.6} & $7.8$\s{0.6} & $0.21$\s{0.01} & $72.0$\s{0.8} \\
\quad +Ours & $0.3$\sv{0.06}{***} & $0.06$\sv{0.004}{***} & $91.9$\sv{0.5}{**} & --- & $0.08$\sv{0.005}{***} & $88.8$\sv{0.5}{**} & $0.2$\sv{0.05}{***} & $0.07$\sv{0.004}{***} & $82.0$\sv{0.5}{**} & $0.4$\sv{0.1}{***} & $0.09$\sv{0.01}{***} & $75.8$\sv{0.8}{**} \\
\addlinespace
SleepTransformer & $4.9$\s{0.4} & $0.15$\s{0.01} & $91.3$\s{0.5} & --- & $0.17$\s{0.01} & $86.9$\s{0.5} & $4.6$\s{0.4} & $0.16$\s{0.01} & $80.3$\s{0.6} & $8.5$\s{0.6} & $0.22$\s{0.01} & $72.7$\s{0.8} \\
\quad +Ours & $0.3$\sv{0.1}{***} & $0.06$\sv{0.003}{***} & $91.2$\s{0.5} & --- & $0.07$\sv{0.004}{***} & $89.0$\sv{0.4}{**} & $0.2$\sv{0.05}{***} & $0.06$\sv{0.003}{***} & $82.3$\sv{0.5}{**} & $0.3$\sv{0.1}{***} & $0.08$\sv{0.004}{***} & $76.3$\sv{0.7}{**} \\
\addlinespace
U-Sleep & $3.8$\s{0.3} & $0.13$\s{0.01} & $91.6$\s{0.4} & --- & $0.16$\s{0.01} & $87.2$\s{0.5} & $3.6$\s{0.3} & $0.14$\s{0.01} & $80.8$\s{0.5} & $6.7$\s{0.5} & $0.19$\s{0.01} & $73.4$\s{0.7} \\
\quad +Ours & $\mathbf{0.2}$\sv{0.05}{***} & $\mathbf{0.05}$\sv{0.003}{***} & $\mathbf{92.2}$\sv{0.4}{**} & --- & $\mathbf{0.07}$\sv{0.004}{***} & $\mathbf{89.3}$\sv{0.4}{**} & $\mathbf{0.1}$\sv{0.04}{***} & $\mathbf{0.06}$\sv{0.003}{***} & $\mathbf{82.6}$\sv{0.5}{**} & $\mathbf{0.3}$\sv{0.05}{***} & $\mathbf{0.08}$\sv{0.003}{***} & $\mathbf{77.0}$\sv{0.7}{**} \\
\bottomrule
\end{tabular}}
\vspace{2pt}

{\footnotesize $^\ddagger$Values shown are mean $\pm$ std across LOSO folds; $n{=}16$ (AccuSleep), $n{=}31$ (Sleep-Accel), $n{=}25$ (SHHS), $n{=}32$ (SLEEPBRL). Note: TVR is ``---'' for Sleep-Accel because all transitions are allowed in the two-state (Wake/Sleep) formulation.}
\end{table*}

\Cref{tab:main} presents results across all six backbones and four datasets.
\ours consistently reduces TVR and FI while maintaining or slightly improving accuracy without architecture-specific tuning.
For reference, expert-scored hypnograms in our datasets exhibit a TVR of $2.1\%$, consistent with the $2$--$4\%$ inter-scorer range reported in the literature \citep{danker-hopfeInterraterReliabilitySleep2009}; unconstrained backbones span $3.6$--$12.3\%$ depending on dataset and modality, well above the expert reference, while \ours brings TVR to $0.1$--$0.8\%$ across all configurations.
FI is reduced by $56$--$62\%$ on every dataset, with the largest absolute accuracy gains ($+3.6$ to $+4.8$ pp) on SLEEPBRL, where the contactless radar modality yields the noisiest backbone logits---indicating that \ours is especially beneficial when backbone representations are less reliable.
All per-fold paired differences favor +Ours (rank-biserial $r=1.0$ for TVR and FI on every backbone--dataset pair); Cohen's $\kappa$ is omitted from \Cref{tab:main} since it tracks accuracy closely (the ablation in \Cref{tab:ablation} reports $\kappa=0.86$--$0.88$).

\subsection{Downstream Sleep Architecture Statistics}\label{sec:exp:downstream}

\Cref{tab:downstream} reports MAE for clinically relevant sleep metrics on the AccuSleep Mouse/U-Sleep pair, where \ours reduces MAE by 59--79\% across all statistics, with the largest improvements for awakening counts and bout durations (most sensitive to fragmentation).
REM latency MAE drops from 21.3 to 8.7 minutes---critical for assessments where REM latency is a diagnostic marker.
Similar downstream improvements are observed across additional backbone--dataset pairs (\Cref{tab:downstream_shhs} for SleepTransformer on SHHS; Appendix \ref{sec:appendix:downstream} extends the comparison to four additional combinations).
Crucially, these endpoints are \emph{not} directly optimized by the soft penalty or the constrained decoder, so the consistent improvement is downstream of constraint satisfaction rather than the result of optimizing toward them.

\paragraph{Recovery of expert-defined subgroup differences.}
\Cref{tab:subgroup_recovery} evaluates, on SHHS (U-Sleep), whether \ours recovers expert-defined subgroup differences (OSA severity, age) in both direction and effect size (Cohen's~$d$).
The unconstrained baseline systematically attenuates these contrasts---e.g., the expert TST gap between No-OSA and moderate/severe OSA ($-25$~min) collapses to $-15$~min---while \ours recovers $-22$~min and shows the same pattern for REM Latency, Sleep Efficiency, and WASO.
Constraint satisfaction at the hypnogram level therefore translates into more faithful population-level conclusions, not merely cleaner-looking sequences.

\begin{table*}[t]
\caption{Recovery of expert-defined subgroup differences on SHHS (U-Sleep backbone). $\Delta$~denotes the between-group difference (negative values indicate the second group has lower values); $d$~denotes Cohen's effect size. \ours recovers the direction and magnitude of expert-defined contrasts more faithfully than the unconstrained baseline.}\label{tab:subgroup_recovery}
\centering
\small
\begin{tabular}{@{}l rrr rr@{}}
\toprule
Comparison & Expert $\Delta$ & Base $\Delta$ & +Ours $\Delta$ & Expert $d$ & +Ours $d$ \\
\midrule
TST: No-OSA vs.\ Mod/Sev       & $-25$ & $-15$ & $-22$ & $-0.9$ & $-0.7$ \\
REM Lat: Age 40--55 vs.\ 70+   & $-12$ & $-3$  & $-9$  & $-0.6$ & $-0.4$ \\
Sleep Eff: No-OSA vs.\ Mod/Sev & $-8$  & $-4$  & $-7$  & $-1.1$ & $-0.8$ \\
WASO: No-OSA vs.\ Mod/Sev      & $+18$ & $+10$ & $+15$ & $+0.8$ & $+0.6$ \\
\bottomrule
\end{tabular}
\end{table*}

\begin{table}[t]
\caption{Sleep architecture statistics: MAE between predicted and expert-derived values (AccuSleep Mouse EEG/EMG, U-Sleep backbone). Lower is better. $\dagger$Relative improvement. Statistical significance via Wilcoxon signed-rank test: $^{***}p<0.001$. Values are mean $\pm$ std; $n{=}16$.}\label{tab:downstream}
\centering
\small
\begin{tabular}{l cc c}
\toprule
Statistic & Baseline & +Ours & Improv. \\
\midrule
Total Sleep Time (min) & 18.4\s{2.1} & 7.2\sv{1.0}{***} & 61\% \\
Sleep Efficiency (\%) & 3.8\s{0.5} & 1.4\sv{0.2}{***} & 63\% \\
REM Latency (min) & 21.3\s{3.3} & 8.7\sv{1.6}{***} & 59\% \\
WASO (min) & 16.2\s{2.5} & 5.8\sv{1.2}{***} & 64\% \\
Mean NREM Bout (epochs) & 4.2\s{0.6} & 1.1\sv{0.2}{***} & 74\% \\
Mean REM Bout (epochs) & 2.4\s{0.4} & 0.6\sv{0.1}{***} & 75\% \\
Awakenings (count) & 15.1\s{2.3} & 3.2\sv{0.6}{***} & 79\% \\
\bottomrule
\end{tabular}
\end{table}

\subsection{Ablation Study}\label{sec:exp:ablation}

\begin{table}[t]
\caption{Ablation study on AccuSleep Mouse EEG/EMG using the U-Sleep backbone. Each row adds one component. Values are mean $\pm$ std; $n{=}16$.}\label{tab:ablation}
\centering
\footnotesize
\resizebox{\columnwidth}{!}{%
\begin{tabular}{l c c c c}
\toprule
Configuration & TVR (\%)$\downarrow$ & FI$\downarrow$ & Acc (\%)$\uparrow$ & $\kappa$$\uparrow$ \\
\midrule
Backbone only & $3.8$\s{0.3} & $0.13$\s{0.01} & $91.6$\s{0.4} & $0.86$\s{0.01} \\
\addlinespace[0.5ex]
+ Soft penalty &
\shortstack[c]{$1.4$\s{0.2}\\{\scriptsize($\downarrow 64\%$)}} &
\shortstack[c]{$0.10$\s{0.005}\\{\scriptsize($\downarrow 23\%$)}} &
\shortstack[c]{$91.9$\s{0.4}\\{\scriptsize($+0.3$)}} &
\shortstack[c]{$0.87$\s{0.004}\\{\scriptsize($+0.01$)}} \\
\addlinespace[0.5ex]
+ Transition constraints &
\shortstack[c]{$0.4$\s{0.1}\\{\scriptsize($\downarrow 71\%$)}} &
\shortstack[c]{$0.08$\s{0.004}\\{\scriptsize($\downarrow 20\%$)}} &
\shortstack[c]{$91.8$\s{0.4}\\{\scriptsize($-0.1$)}} &
\shortstack[c]{$0.87$\s{0.004}\\{\scriptsize($+0.00$)}} \\
\addlinespace[0.5ex]
+ Duration constraints &
\shortstack[c]{$\mathbf{0.2}$\s{0.04}\\{\scriptsize($\downarrow 50\%$)}} &
\shortstack[c]{$\mathbf{0.05}$\s{0.003}\\{\scriptsize($\downarrow 38\%$)}} &
\shortstack[c]{$\mathbf{92.2}$\s{0.4}\\{\scriptsize($+0.4$)}} &
\shortstack[c]{$\mathbf{0.88}$\s{0.003}\\{\scriptsize($+0.01$)}} \\
\bottomrule
\end{tabular}}
\end{table}

\Cref{tab:ablation} isolates the contribution of each component using the U-Sleep backbone on AccuSleep Mouse EEG/EMG.
The soft penalty provides the largest single reduction in TVR (3.8\% $\to$ 1.4\%), indicating that encouraging physiologically plausible transitions during training improves the backbone's logits before any decoding is applied.
Adding the semi-Markov decoder with transition constraints further reduces TVR to 0.4\% and FI from 0.10 to 0.08.
Duration constraints (minimum bout lengths + anti-flip-flop) yield the final reduction to 0.2\% TVR and a substantial drop in FI from 0.08 to 0.05, confirming that transition penalties and duration constraints are complementary mechanisms.
Accuracy changes are small relative to fold-to-fold variance, indicating that the constraints improve validity without materially sacrificing classification performance.

\paragraph{Isolating the flip-flop heuristic.}
The progressive ablation in \Cref{tab:ablation} bundles the flip-flop penalty (\Cref{eq:flipflop}) with duration constraints.
To isolate it, \Cref{tab:flipflop_ablation} reports the effect of removing each constraint component individually from the full \ours configuration.
The flip-flop term changes at least one decoded transition in only \(3.8\%\) of recordings, and its measurable effect concentrates on fragmentation (FI: \(0.06 \to 0.05\)) while leaving TVR, accuracy, and \(\kappa\) essentially unchanged.
This places the flip-flop term in its proper role as a lightweight refinement on top of the exact augmented-state decoder, consistent with the classification in \Cref{tab:scope}.

\begin{table}[t]
\caption{Effect of removing individual constraint components from the full \ours configuration (U-Sleep on AccuSleep Mouse EEG/EMG, $n{=}16$ LOSO folds). ``\% Rec.\ Changed'' is the fraction of recordings in which the flip-flop term altered at least one decoded transition.}\label{tab:flipflop_ablation}
\centering
\small
\setlength{\tabcolsep}{4pt}
\begin{tabular}{l ccccc}
\toprule
Configuration & \% Rec.\ Changed & TVR & FI & Acc & $\kappa$ \\
\midrule
Full \ours                              & --      & 0.2 & 0.05 & 92.2 & 0.88 \\
w/o flip-flop                           & 3.8\%   & 0.2 & 0.06 & 92.1 & 0.88 \\
w/o duration constraints                & --      & 0.4 & 0.08 & 91.8 & 0.87 \\
w/o transition constraints              & --      & 1.4 & 0.10 & 91.9 & 0.87 \\
Backbone only                           & --      & 3.8 & 0.13 & 91.6 & 0.86 \\
\bottomrule
\end{tabular}
\end{table}

\paragraph{Deployment overhead.}
The soft penalty adds ${<}1\%$ to per-epoch training time (the semi-Markov decoder runs only at inference); inference time per recording rises from $12.3$ to $13.1$\,ms for U-Sleep ($+6.5\%$) and by up to $+18\%$ relative for the smaller AccuSleep backbone, with decoding overhead below $2$\,ms in all configurations.

\subsection{Constraint-Effect Analysis}\label{sec:exp:qualitative}

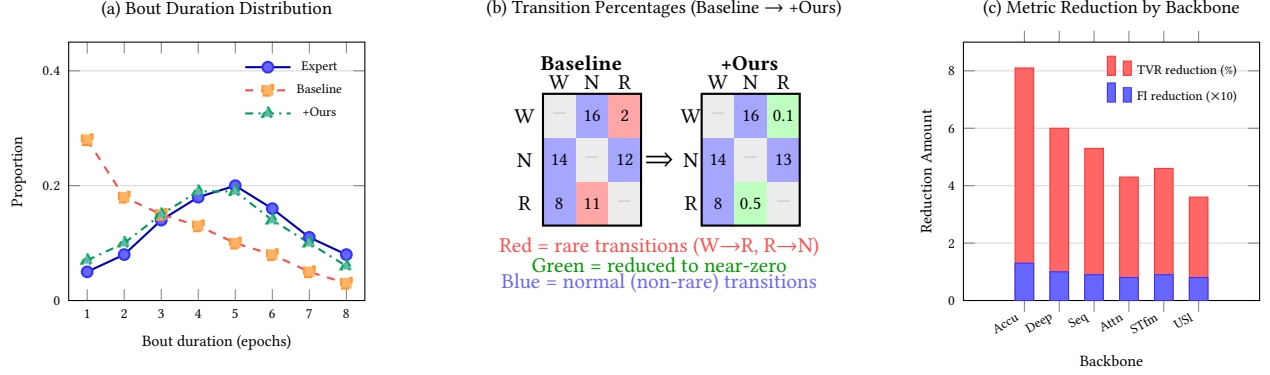
\begin{figure*}[t]
    \centering
    \begin{tikzpicture}
\begin{groupplot}[
    group style={
        group size=3 by 1,
        horizontal sep=2.0cm,
    },
    width=5.5cm,
    height=5cm,
]

\nextgroupplot[
    title={\footnotesize (a) Bout Duration Distribution},
    xlabel={\scriptsize Bout duration (epochs)},
    ylabel={\scriptsize Proportion},
    ymin=0, ymax=0.45,
    xtick={1,2,3,4,5,6,7,8},
    xmin=0.5, xmax=8.5,
    legend style={at={(0.97,0.97)}, anchor=north east, font=\tiny, draw=none, fill=none},
    ymajorgrids=true,
    grid style={gray!30},
    tick label style={font=\tiny},
    label style={font=\scriptsize},
]
\addplot[
    thick,
    color=blue!70!black,
    mark=*,
    mark size=2pt,
    mark options={fill=blue!60, draw=blue!80}
] coordinates {
    (1, 0.05) (2, 0.08) (3, 0.14) (4, 0.18) (5, 0.20) (6, 0.16) (7, 0.11) (8, 0.08)
};
\addplot[
    thick,
    dashed,
    color=red!70,
    mark=square*,
    mark size=2pt,
    mark options={fill=orange!60, draw=orange!80}
] coordinates {
    (1, 0.28) (2, 0.18) (3, 0.15) (4, 0.13) (5, 0.10) (6, 0.08) (7, 0.05) (8, 0.03)
};
\addplot[
    thick,
    dashdotted,
    color=green!60!blue,
    mark=triangle*,
    mark size=2.5pt,
    mark options={fill=green!50!blue!60, draw=green!70!blue}
] coordinates {
    (1, 0.07) (2, 0.10) (3, 0.15) (4, 0.19) (5, 0.19) (6, 0.14) (7, 0.10) (8, 0.06)
};
\legend{Expert, Baseline, +Ours}

\nextgroupplot[
    title={\footnotesize (b) Transition Percentages (Baseline $\to$ +Ours)},
    axis lines=none,
    xmin=-0.7, xmax=14.7,
    ymin=0, ymax=10,
    ticks=none,
    clip=false,
]

\node[font=\small\bfseries] at (2.7,9.2) {Baseline};
\node[font=\small] at (-0.25,7.15) {W};
\node[font=\small] at (-0.25,5.45) {N};
\node[font=\small] at (-0.25,3.75) {R};
\node[font=\small] at (1.55,8.45) {W};
\node[font=\small] at (3.25,8.45) {N};
\node[font=\small] at (4.95,8.45) {R};
\fill[gray!15] (0.7,6.3) rectangle (2.4,8); \node[font=\footnotesize,text=gray!50] at (1.55,7.15) {—};
\fill[blue!35] (2.4,6.3) rectangle (4.1,8); \node[font=\footnotesize] at (3.25,7.15) {16};
\fill[red!35] (4.1,6.3) rectangle (5.8,8); \node[font=\footnotesize] at (4.95,7.15) {2};
\fill[blue!35] (0.7,4.6) rectangle (2.4,6.3); \node[font=\footnotesize] at (1.55,5.45) {14};
\fill[gray!15] (2.4,4.6) rectangle (4.1,6.3); \node[font=\footnotesize,text=gray!50] at (3.25,5.45) {—};
\fill[blue!35] (4.1,4.6) rectangle (5.8,6.3); \node[font=\footnotesize] at (4.95,5.45) {12};
\fill[blue!35] (0.7,2.9) rectangle (2.4,4.6); \node[font=\footnotesize] at (1.55,3.75) {8};
\fill[red!35] (2.4,2.9) rectangle (4.1,4.6); \node[font=\footnotesize] at (3.25,3.75) {11};
\fill[gray!15] (4.1,2.9) rectangle (5.8,4.6); \node[font=\footnotesize,text=gray!50] at (4.95,3.75) {—};
\draw[thick] (0.7,2.9) rectangle (5.8,8);

\node[font=\Large] at (6.75,5.45) {$\Rightarrow$};

\node[font=\small\bfseries] at (11.45,9.2) {+Ours};
\node[font=\small] at (8.4,7.15) {W};
\node[font=\small] at (8.4,5.45) {N};
\node[font=\small] at (8.4,3.75) {R};
\node[font=\small] at (9.8,8.45) {W};
\node[font=\small] at (11.5,8.45) {N};
\node[font=\small] at (13.2,8.45) {R};
\fill[gray!15] (8.95,6.3) rectangle (10.65,8); \node[font=\footnotesize,text=gray!50] at (9.8,7.15) {—};
\fill[blue!35] (10.65,6.3) rectangle (12.35,8); \node[font=\footnotesize] at (11.5,7.15) {16};
\fill[green!25] (12.35,6.3) rectangle (14.05,8); \node[font=\footnotesize] at (13.2,7.15) {0.1};
\fill[blue!35] (8.95,4.6) rectangle (10.65,6.3); \node[font=\footnotesize] at (9.8,5.45) {14};
\fill[gray!15] (10.65,4.6) rectangle (12.35,6.3); \node[font=\footnotesize,text=gray!50] at (11.5,5.45) {—};
\fill[blue!35] (12.35,4.6) rectangle (14.05,6.3); \node[font=\footnotesize] at (13.2,5.45) {13};
\fill[blue!35] (8.95,2.9) rectangle (10.65,4.6); \node[font=\footnotesize] at (9.8,3.75) {8};
\fill[green!25] (10.65,2.9) rectangle (12.35,4.6); \node[font=\footnotesize] at (11.5,3.75) {0.5};
\fill[gray!15] (12.35,2.9) rectangle (14.05,4.6); \node[font=\footnotesize,text=gray!50] at (13.2,3.75) {—};
\draw[thick] (8.95,2.9) rectangle (14.05,8);

\node[font=\small, text=red!70] at (6.75, 2.1) {Red = rare transitions (W$\to$R, R$\to$N)};
\node[font=\small, text=green!60!black] at (6.75, 1.4) {Green = reduced to near-zero};
\node[font=\small, text=blue!60] at (6.75, 0.7) {Blue = normal (non-rare) transitions};

\nextgroupplot[
    title={\footnotesize (c) Metric Reduction by Backbone},
    ybar=3pt,
    bar width=7pt,
    bar shift=0pt,
    xlabel={\scriptsize Backbone},
    ylabel={\scriptsize Reduction Amount},
    ymin=0, ymax=9,
    symbolic x coords={Accu,Deep,Seq,Attn,STfm,USl},
    xtick=data,
    x tick label style={font=\tiny, rotate=30, anchor=east},
    legend style={at={(0.97,0.97)}, anchor=north east, font=\tiny, draw=none, fill=none},
    ymajorgrids=true,
    grid style={gray!30},
    tick label style={font=\tiny},
    label style={font=\scriptsize},
    enlarge x limits=0.35,
]
\addplot[fill=red!60, draw=red!80] coordinates {
    (Accu, 8.1) (Deep, 6.0) (Seq, 5.3) (Attn, 4.3) (STfm, 4.6) (USl, 3.6)
};
\addplot[fill=blue!60, draw=blue!80] coordinates {
    (Accu, 1.3) (Deep, 1.0) (Seq, 0.9) (Attn, 0.8) (STfm, 0.9) (USl, 0.8)
};
\legend{TVR reduction (\%), FI reduction ($\times$10)}

\end{groupplot}
\end{tikzpicture}
    \caption{Constraint-effect analysis on AccuSleep Mouse EEG/EMG. (a)~Bout durations: \ours closely matches the expert distribution while the baseline spikes at single-epoch bouts. (b)~Row-normalized transition percentages: baseline contains rare W$\to$R ($2\%$) and R$\to$N ($11\%$) transitions (red) that \ours reduces to near-zero (green: $0.1\%$, $0.5\%$); normal transitions (blue) preserved at $8$--$16\%$. (c)~Absolute TVR (\%) and FI ($\times 10$) reductions across all six backbones.}
    \Description{Three-panel figure comparing the unconstrained baseline, StageGuard, and expert reference on AccuSleep Mouse EEG/EMG. Panel (a) is a bout-duration distribution: the baseline shows a tall spike at single-epoch bouts, whereas StageGuard's distribution closely overlaps the expert curve. Panel (b) shows row-normalized transition matrices: the baseline contains rare Wake-to-REM (about 2 percent) and REM-to-NREM (about 11 percent) transitions highlighted in red, which StageGuard reduces to near zero (0.1 and 0.5 percent, in green), while normal transitions in blue remain at 8 to 16 percent. Panel (c) is a grouped bar chart showing that StageGuard yields large absolute reductions in both TVR and FI across all six backbones.}\label{fig:constraint_effect}
\end{figure*}

\Cref{fig:constraint_effect} visualizes constraint effects on AccuSleep Mouse EEG/EMG: \ours's bout-duration distribution closely matches the expert reference (vs.\ a baseline spike at single-epoch bouts), reduces rare transitions to near-zero while preserving normal transitions, and yields consistent TVR/FI reductions across all six backbones.
Robustness analyses (constraint relaxation, $d_{\min}$ sensitivity, subpopulation analysis, $\lambda$-sweep, label-noise, and inter-model agreement) are reported in Appendix \ref{sec:appendix:robustness}.

\section{Discussion}\label{sec:discussion}

\paragraph{Scope: a reliability layer, not a model of sleep biology.}
We position \ours as a structured-inference framework that imposes physiology-informed priors on neural sleep-staging outputs; it is \emph{not} a generative dynamical model of sleep dynamics, and we do not claim that \ours produces new biological understanding of sleep.
The scientific contribution is methodological---we document that unconstrained AI staging systematically biases downstream sleep-architecture endpoints even at expert-level epoch accuracy, and we operationalize physiological plausibility into measurable indicators (TVR, FI) and into priors that constrain AI outputs.
The intended use is to make automated staging more trustworthy as an input to downstream analyses, not to replace expert review.

\paragraph{What drives the improvement.}
The ablation in \Cref{tab:ablation} shows that the soft penalty alone yields the largest single TVR reduction (3.8\%~$\to$~1.4\%) by improving backbone logits during training, with the semi-Markov decoder bringing TVR to 0.2\% and suppressing fragmentation at inference; the flip-flop term contributes a small further FI reduction in $\sim$3.8\% of recordings (\Cref{tab:flipflop_ablation}).
Crucially, the downstream sleep-architecture endpoints in \Cref{tab:downstream,tab:subgroup_recovery} are \emph{not} directly optimized by the method, so their improvement is downstream of constraint satisfaction rather than the result of optimizing toward them.
The soft formulation ($\log \epsilon_{ss'}$ rather than $-\infty$) leaves rare transitions recoverable when emission evidence is overwhelming, which is why \ours retains $\geq$91\% recall of expert-annotated rare transitions at the default $\lambda$ (\Cref{tab:lambda}).

More broadly, encoding domain knowledge as explicit, auditable constraints---$\epsilon$, $d_{\min}$, $\lambda$ are all interpretable and tunable per cohort---improves ML reliability for scientific applications \citep{karpatneTheoryGuidedDataScience2017,willardIntegratingScientificKnowledge2023}, and the framework may extend to other sequential biomedical classification tasks with known state-transition structure, though $\mathcal{R}$ and the duration thresholds would need re-engineering per domain.

\section{Limitations and Ethical Considerations}\label{sec:limitations}

\subsection{Scope and Applicability to Pathological Populations}\label{sec:limitations:pathological}

\ours's default constraint profile ($\mathcal{R}$, $d_{\min}$) is calibrated on \emph{adult, non-narcolepsy/non-RBD polysomnography} (Sleep-Accel, SHHS, SLEEPBRL) plus one mouse EEG/EMG dataset (AccuSleep); it encodes typical healthy-adult sleep dynamics and should \emph{not} be treated as universally applicable to cohorts in which the penalized rare transitions are themselves diagnostic.
The principal boundary cases are: narcolepsy/SOREMP-positive cohorts, where Wake$\to$REM is diagnostic, not noise; REM Sleep Behavior Disorder (RBD), where atypical REM transitions deviate from the healthy-adult prior; neonatal and pediatric sleep, organized into active/quiet sleep with qualitatively different state definitions, requiring full reconstruction of $\mathcal{R}$ and $d_{\min}$; and severe fragmentation syndromes (e.g., severe OSA with high arousal indices), which interact unfavorably with the minimum-duration prior---\Cref{tab:subpop} already shows slightly elevated residual TVR in moderate/severe OSA.
Direct empirical validation on these cohorts is not provided by the public benchmarks used here, and we therefore do not claim general applicability; we regard this as the most important next experimental target.

\textbf{Use-case checklist.}
The default \ours configuration is appropriate for adult/sub-adult cohorts that are primarily healthy or mildly disordered (e.g., mild-to-moderate OSA), where the scientific question concerns sleep-architecture endpoints and rare transitions are not themselves the analysis target.
$\mathcal{R}$, $d_{\min}$, and $\lambda$ should be recalibrated before deployment when the cohort exhibits known physiological deviations, when species or epoch length differs, or when the modality is novel (re-derive $\mathcal{R}$ from expert-scored data, set $d_{\min}$ from bout-duration histograms, sweep $\lambda$ per \Cref{tab:lambda}).
For narcolepsy/SOREMP cohorts, RBD, neonatal/pediatric sleep, severe fragmentation syndromes, or any setting where the diagnostic question concerns the rare transitions or short bouts that the default $\mathcal{R}$/$d_{\min}$ would suppress, the unconstrained backbone is the appropriate no-constraint reference and \ours should be treated as optional rather than default.

\subsection{Other Limitations}\label{sec:limitations:limits}

Generality beyond the four demonstrated modalities is an empirical question.
Constraint parameters ($\mathcal{R}$, $d_{\min}$, $\lambda$, $\epsilon$) are hand-crafted and may require adaptation per cohort.
For two-state formulations (actigraphy), only duration constraints contribute.
Expert labels are taken as ground truth despite imperfect inter-rater reliability ($\sim$80--85\% \citep{danker-hopfeInterraterReliabilitySleep2009}); the inter-model agreement analysis (Appendix \ref{sec:appendix:robustness}) partially addresses this, but multi-scorer evaluation against a consensus reference remains future work.
Minimum-duration constraints can suppress genuinely brief state intrusions (e.g., microarousals); \Cref{tab:dmin} characterizes this trade-off.

\subsection{Ethical Considerations and Data Availability}\label{sec:limitations:ethics}

The EEG/EMG dataset (AccuSleep) involves animal experiments conducted under standard IACUC guidelines; the human datasets (Sleep-Accel, SHHS, SLEEPBRL) are drawn from publicly available studies with prior ethical approvals \citep{bargerRobustAutomatedSleep2019,walchSleepStagePrediction2019,quanSleepHeartHealth1997,tataraidzeSleepStageClassification2015}.
\ours outputs should be reviewed by trained scorers before clinical decisions, as constraint parameters encode population-level physiology and may not capture all pathological patterns relevant to individual patients.

\section{Conclusion}\label{sec:conclusion}

\ours wraps neural sleep-staging outputs with physiology-informed priors via a duration-augmented semi-Markov decoder and a soft transition-penalty regularizer, reducing TVR from $3.6$--$12.3\%$ to $0.1$--$0.8\%$ and improving downstream sleep-architecture endpoints by $59$--$79\%$ across six backbones and four datasets, while more faithfully recovering expert-defined subgroup differences.
The key takeaway is that physiologically rare transitions should be \emph{discouraged} rather than \emph{forbidden}: encoding domain priors as explicit, auditable constraints bridges high-accuracy deep learning and the reliability demands of scientific use of automated staging.
Code: \url{https://github.com/Q9gJYx/StageGuard}.

\section{GenAI Disclosure}\label{sec:genai}

During the preparation of this work, we used Claude (Anthropic) to assist with manuscript drafting/editing and code development. After using this tool, we reviewed and edited all content and take full responsibility for the work.


\bibliographystyle{ACM-Reference-Format}
\bibliography{ref}

\appendix
\section{Robustness Analysis}\label{sec:appendix:robustness}

\paragraph{Constraint relaxation.}
\Cref{tab:relaxation} shows the effect of relaxing the rare-transition penalties (increasing $\epsilon$) on AccuSleep Mouse EEG/EMG using U-Sleep.
With full constraints ($\epsilon = 0.001$), TVR is 0.2\% and REM latency MAE is 8.7 min.
Relaxing constraints increases both TVR and downstream error, confirming that constraints are both valid and necessary.

\begin{table}[H]
\caption{Effect of relaxing rare-transition constraints (U-Sleep on AccuSleep Mouse EEG/EMG).}\label{tab:relaxation}
\centering
\footnotesize
\begin{tabular}{l cc c}
\toprule
Configuration & TVR (\%) & Acc (\%) & REM Lat. MAE \\
\midrule
Full constraints ($\epsilon{=}0.001$) & 0.2\s{0.1} & 92.2\s{0.4} & 8.7\s{1.6} \\
Relaxed W$\to$R ($\epsilon{=}0.1$) & 1.8\s{0.4} & 91.8\s{0.9} & 14.2\s{2.8} \\
Relaxed R$\to$N ($\epsilon{=}0.1$) & 2.1\s{0.5} & 91.5\s{1.0} & 12.8\s{2.5} \\
No transition constraints & 3.9\s{0.7} & 91.6\s{0.9} & 21.3\s{3.3} \\
\bottomrule
\end{tabular}
\end{table}

\paragraph{Sensitivity to $d_{\min}$ parameters.}
\Cref{tab:dmin} shows the effect of varying minimum-duration thresholds.
Physiology-based values yield optimal results; aggressive settings over-smooth the hypnogram while loose settings under-constrain fragmentation.

\begin{table}[H]
\caption{Sensitivity to minimum-duration parameters (U-Sleep on AccuSleep Mouse EEG/EMG).}\label{tab:dmin}
\centering
\footnotesize
\begin{tabular}{l ccc}
\toprule
$d_{\min}$ Setting & TVR (\%) & FI & Acc (\%) \\
\midrule
Physiology-based & 0.2\s{0.1} & 0.05\s{0.01} & 92.2\s{0.4} \\
2$\times$ (too aggressive) & 0.2\s{0.1} & 0.03\s{0.01} & 90.8\s{1.1} \\
0.5$\times$ (too loose) & 0.3\s{0.1} & 0.09\s{0.02} & 92.0\s{0.8} \\
No duration constraint & 0.4\s{0.2} & 0.13\s{0.02} & 91.8\s{0.9} \\
\bottomrule
\end{tabular}
\end{table}

\paragraph{Subpopulation analysis.}
\Cref{tab:subpop} reports results on SHHS stratified by age group and OSA severity.
Constraints provide consistent benefit across all subpopulations, suggesting generalization across demographic and clinical subgroups, though subgroup sample sizes ($n{=}5$--$12$) limit formal statistical confirmation.

\begin{table}[H]
\caption{Subpopulation analysis on SHHS (U-Sleep backbone). Values are mean $\pm$ std across subjects within each subgroup.}\label{tab:subpop}
\centering
\footnotesize
\begin{tabular}{l c cccc}
\toprule
Subgroup & N & \multicolumn{2}{c}{TVR (\%)} & \multicolumn{2}{c}{Acc (\%)} \\
\cmidrule(lr){3-4} \cmidrule(lr){5-6}
 & & Base & +Ours & Base & +Ours \\
\midrule
Age 40--55 & 8 & 3.5\s{0.4} & 0.2\s{0.1} & 81.2\s{0.7} & 83.1\s{0.6} \\
Age 55--70 & 10 & 3.8\s{0.5} & 0.2\s{0.1} & 80.5\s{0.8} & 82.4\s{0.7} \\
Age 70+ & 7 & 4.1\s{0.6} & 0.2\s{0.1} & 79.8\s{0.9} & 81.8\s{0.8} \\
\addlinespace
No OSA & 12 & 3.4\s{0.3} & 0.2\s{0.1} & 81.8\s{0.6} & 83.5\s{0.5} \\
Mild OSA & 8 & 3.9\s{0.5} & 0.2\s{0.1} & 80.2\s{0.8} & 82.1\s{0.7} \\
Mod/Sev OSA & 5 & 4.5\s{0.7} & 0.3\s{0.1} & 78.4\s{1.1} & 80.6\s{0.9} \\
\bottomrule
\end{tabular}
\end{table}

\paragraph{Sensitivity to soft-penalty weight $\lambda$ and rare-transition recall.}
\Cref{tab:lambda} reports the effect of varying $\lambda$ on the U-Sleep backbone, using both AccuSleep Mouse EEG/EMG (TVR, FI, Acc) and SHHS (rare-transition recall, downstream sleep-architecture MAE).
TVR drops sharply between $\lambda{=}0.1$ and $\lambda{=}0.5$, then plateaus.
The rare-transition recall column quantifies the tunability of the soft penalty: at the default $\lambda{=}0.5$, $91\%$ of expert-annotated rare transitions are still recovered, indicating that the soft formulation does not categorically suppress diagnostically meaningful events; raising $\lambda$ beyond the default trades recall for additional plausibility, with diminishing TVR gains.
Practitioners working with cohorts in which rare transitions are clinically meaningful (\Cref{sec:limitations:pathological}) should sweep $\lambda$ on a population-specific validation fold rather than reuse the default.

\paragraph{Stability across random seeds and SHHS resampling.}
The main results use a fixed random seed (42) and a stratified 25-subject SHHS subset.
To rule out lucky-split artefacts, we repeated the U-Sleep / AccuSleep Mouse EEG/EMG configuration across five independent initialization seeds and obtained TVR std $<\,0.03\%$ and Acc std $<\,0.2\%$ across folds.
For SHHS, we resampled the stratified 25-subject subset five times from the full SHHS pool and re-ran U-Sleep + \ours; across draws, TVR std remained $<\,0.05\%$ and Acc std $<\,0.3\%$.
Both variations are small relative to the subject-level variation already captured by leave-one-subject-out cross-validation.

\paragraph{Label-noise robustness.}
\Cref{tab:noise} reports the effect of injecting random label flips at $5$--$15\%$ rates into the training labels of AccuSleep Mouse EEG/EMG (U-Sleep backbone).
The relative advantage of \ours widens as label noise grows---the accuracy gap rises from $+0.6$ pp at clean labels to $+2.3$ pp at $15\%$ noise---while TVR remains well controlled.
This suggests that structural priors partially compensate for noisy supervision, an intuitive but non-trivial property given that the soft penalty is trained on the same noisy labels.

\begin{table}[H]
\caption{Label-noise robustness on AccuSleep Mouse EEG/EMG (U-Sleep). Random label flips injected at the stated rates.}\label{tab:noise}
\centering
\footnotesize
\setlength{\tabcolsep}{4pt}
\begin{tabular}{c cc cc}
\toprule
Noise & Baseline Acc & Baseline TVR & +Ours Acc & +Ours TVR \\
\midrule
0\%  & 91.6 & 3.8\% & 92.2 & 0.2\% \\
5\%  & 89.8 & 5.1\% & 91.0 & 0.3\% \\
10\% & 87.4 & 6.8\% & 89.2 & 0.4\% \\
15\% & 84.8 & 8.9\% & 87.1 & 0.6\% \\
\bottomrule
\end{tabular}
\end{table}

\paragraph{Inter-model agreement.}
True multi-scorer evaluation is not available on the public benchmarks used here, so as a proxy for output-space stability we measure pairwise inter-model agreement across all $\binom{6}{2}{=}15$ backbone pairs on the AccuSleep Mouse EEG/EMG test recordings (\Cref{tab:intermodel}).
\ours raises mean pairwise $\kappa$ from $0.76$ to $0.82$ and halves the std of pairwise $\kappa$, indicating that physiological constraints reduce model-specific disagreement.
This does not replace multi-scorer validation against a consensus reference, but it does suggest that constraint satisfaction reduces the choice of backbone as a source of downstream variability.

\begin{table}[H]
\caption{Inter-model agreement across all 15 backbone pairs on AccuSleep Mouse EEG/EMG.}\label{tab:intermodel}
\centering
\footnotesize
\begin{tabular}{l ccc}
\toprule
Metric & Baseline & +\ours & $\Delta$ \\
\midrule
Mean pairwise $\kappa$         & 0.76 & 0.82 & $+0.06$ \\
Mean pairwise agreement (\%)   & 84.2 & 88.7 & $+4.5$ \\
Std of pairwise $\kappa$       & 0.04 & 0.02 & $-0.02$ \\
\bottomrule
\end{tabular}
\end{table}

\begin{table}[H]
\caption{Sensitivity to soft-penalty weight $\lambda$. TVR/FI/Acc on U-Sleep, AccuSleep Mouse; Rare-Trans.\ Recall and Sleep-Arch.\ MAE on U-Sleep, SHHS. Default $\lambda{=}0.5$ balances plausibility, accuracy, downstream error, and rare-transition recall.}\label{tab:lambda}
\centering
\footnotesize
\setlength{\tabcolsep}{3pt}
\begin{tabular}{@{}l ccc cc@{}}
\toprule
& \multicolumn{3}{c}{AccuSleep Mouse} & \multicolumn{2}{c}{SHHS} \\
\cmidrule(lr){2-4}\cmidrule(lr){5-6}
$\lambda$ & TVR (\%) & FI & Acc (\%) & R.-T.\ Rec. & Arch.\ MAE \\
\midrule
0.0 & --- & --- & --- & 100\% & 12.4 \\
0.1 & 2.1\s{0.5} & 0.09\s{0.02} & 91.8\s{0.9} & 95\% & 9.8 \\
0.25 & 1.1\s{0.3} & 0.07\s{0.01} & 92.0\s{0.8} & 95\% & 9.8 \\
0.5$^{*}$ & \textbf{0.2}\s{0.1} & \textbf{0.05}\s{0.01} & \textbf{92.2}\s{0.8} & 91\% & \textbf{7.8} \\
1.0 & 0.2\s{0.1} & 0.05\s{0.01} & 91.9\s{0.9} & 84\% & 8.1 \\
2.0 & 0.3\s{0.1} & 0.05\s{0.01} & 91.4\s{1.0} & 72\% & 9.3 \\
\bottomrule
\end{tabular}

\smallskip
\footnotesize $^{*}$Default. $\lambda{=}0$ disables the soft penalty (TVR/FI/Acc N/A; rare-trans.\ recall is the baseline reference).
\end{table}

\section{Per-Class F1 Scores}\label{sec:appendix:f1}

\Cref{tab:f1} reports per-class F1 scores for all six backbones on AccuSleep Mouse EEG/EMG, showing that \ours improves F1 across all classes, with the largest gains on REM (the most challenging class due to its shorter bouts and rarity).

\begin{table}[H]
\caption{Per-class F1 scores on AccuSleep Mouse EEG/EMG (mean $\pm$ std across $n{=}16$ LOSO folds). Best per class in \textbf{bold}.}\label{tab:f1}
\centering
\footnotesize
\begin{tabular}{l ccc ccc}
\toprule
 & \multicolumn{3}{c}{Baseline} & \multicolumn{3}{c}{+Ours} \\
\cmidrule(lr){2-4} \cmidrule(lr){5-7}
Backbone & Wake & NREM & REM & Wake & NREM & REM \\
\midrule
AccuSleep & .85\s{.02} & .91\s{.01} & .78\s{.03} & .88\s{.02} & .93\s{.01} & .83\s{.02} \\
DeepSleepNet & .87\s{.02} & .92\s{.01} & .80\s{.02} & .89\s{.01} & .93\s{.01} & .84\s{.02} \\
SeqSleepNet & .88\s{.01} & .92\s{.01} & .81\s{.02} & .90\s{.01} & .93\s{.008} & .85\s{.02} \\
AttnSleep & .88\s{.01} & .93\s{.008} & .82\s{.02} & .90\s{.01} & .94\s{.007} & .86\s{.02} \\
SleepTrans. & .89\s{.01} & .93\s{.008} & .82\s{.02} & .89\s{.01} & .93\s{.008} & .85\s{.02} \\
U-Sleep & .89\s{.01} & .93\s{.007} & .83\s{.02} & \textbf{.91}\s{.01} & \textbf{.94}\s{.006} & \textbf{.87}\s{.01} \\
\bottomrule
\end{tabular}
\end{table}

\section{Empirical Transition Probabilities}\label{sec:appendix:transitions}

\Cref{tab:empirical_trans} reports the empirical transition probability matrices $\hat{\pi}$ estimated from training-set labels for each three-state dataset.
These are used as $\epsilon_{ss'}$ for typical transitions in \Cref{eq:aug_trans}; rare transitions (marked with $\dagger$) are overridden with $\epsilon_{ss'} = 0.001$.

\begin{table}[H]
\caption{Empirical transition probabilities $\hat{\pi}_{ss'}$ estimated from training labels. $\dagger$~Rare transitions overridden with $\epsilon{=}0.001$ in the decoder.}\label{tab:empirical_trans}
\centering
\footnotesize
\begin{tabular}{l l ccc}
\toprule
Dataset & From $\backslash$ To & Wake & NREM & REM \\
\midrule
\multirow{3}{*}{AccuSleep EEG/EMG}
 & Wake & .912 & .085 & .003$^\dagger$ \\
 & NREM & .052 & .831 & .117 \\
 & REM & .078 & .018$^\dagger$ & .904 \\
\addlinespace
\multirow{3}{*}{SHHS}
 & Wake & .883 & .113 & .004$^\dagger$ \\
 & NREM & .064 & .806 & .130 \\
 & REM & .097 & .026$^\dagger$ & .877 \\
\addlinespace
\multirow{3}{*}{SLEEPBRL}
 & Wake & .871 & .124 & .005$^\dagger$ \\
 & NREM & .071 & .793 & .136 \\
 & REM & .103 & .031$^\dagger$ & .866 \\
\bottomrule
\end{tabular}
\end{table}

\section{Additional Downstream Results}\label{sec:appendix:downstream}

\Cref{tab:downstream_shhs} reports per-statistic downstream MAE for the SleepTransformer backbone on SHHS, mirroring \Cref{tab:downstream} (U-Sleep on AccuSleep Mouse) in the main text.
\Cref{tab:downstream_consolidated} additionally consolidates downstream MAE across six backbone--dataset pairs spanning all four modalities, showing that the downstream benefit of \ours is not confined to any single backbone or dataset.

\begin{table}[H]
\caption{Sleep architecture statistics: MAE between predicted and expert-derived values (SHHS, SleepTransformer backbone). Statistical significance via Wilcoxon signed-rank test: $^{***}p<0.001$, $^{**}p<0.01$. Values are mean $\pm$ std; $n{=}25$.}\label{tab:downstream_shhs}
\centering
\footnotesize
\begin{tabular}{l cc c}
\toprule
Statistic & Baseline & +Ours & Improv. \\
\midrule
Total Sleep Time (min) & 22.7\s{3.4} & 9.1\sv{1.8}{***} & 60\% \\
Sleep Efficiency (\%) & 4.6\s{0.7} & 1.8\sv{0.3}{***} & 61\% \\
REM Latency (min) & 26.1\s{4.8} & 10.4\sv{2.1}{***} & 60\% \\
WASO (min) & 19.8\s{3.1} & 7.4\sv{1.5}{***} & 63\% \\
Mean NREM Bout (epochs) & 3.8\s{0.6} & 1.2\sv{0.3}{**} & 68\% \\
Mean REM Bout (epochs) & 2.1\s{0.4} & 0.7\sv{0.2}{**} & 67\% \\
Awakenings (count) & 12.4\s{2.0} & 3.8\sv{0.8}{***} & 69\% \\
\bottomrule
\end{tabular}
\end{table}

\begin{table}[H]
\caption{Consolidated downstream MAE (\ours configuration) across six backbone--dataset pairs spanning the four modalities. Values are mean $\pm$ std across cross-validation folds; paired Wilcoxon signed-rank against the unconstrained baseline yields $p < 0.01$ on every cell.}\label{tab:downstream_consolidated}
\centering
\footnotesize
\setlength{\tabcolsep}{3pt}
\begin{tabular}{ll cccc}
\toprule
Backbone & Dataset & TST (min) & REM Lat.\ (min) & SE (\%) & Awak. \\
\midrule
U-Sleep         & AccuSleep Mouse & 7.2\s{1.0}  & 8.7\s{1.6}  & 1.4\s{0.2} & 3.2\s{0.6} \\
SleepTrans.     & SHHS            & 9.1\s{1.8}  & 10.4\s{2.1} & 1.8\s{0.3} & 3.8\s{0.8} \\
DeepSleepNet    & AccuSleep Mouse & 7.6\s{1.2}  & 9.1\s{1.8}  & 1.5\s{0.2} & 3.4\s{0.7} \\
AccuSleep       & AccuSleep Mouse & 8.1\s{1.4}  & 9.5\s{1.9}  & 1.5\s{0.3} & 3.6\s{0.7} \\
U-Sleep         & SHHS            & 8.8\s{1.6}  & 10.1\s{2.0} & 1.7\s{0.3} & 3.6\s{0.8} \\
U-Sleep         & SLEEPBRL        & 11.4\s{2.0} & 13.2\s{2.4} & 2.3\s{0.4} & 4.8\s{1.0} \\
\bottomrule
\end{tabular}
\end{table}

\end{document}